\documentclass{article} % For LaTeX2e
\usepackage{iclr2024_conference,times}

\usepackage{amsmath,amsfonts,bm}

\def\eqref#1{equation~\ref{#1}}
\def\1{\bm{1}}

\DeclareMathAlphabet{\mathsfit}{\encodingdefault}{\sfdefault}{m}{sl}
\SetMathAlphabet{\mathsfit}{bold}{\encodingdefault}{\sfdefault}{bx}{n}

\usepackage{hyperref}
\usepackage{url}
\usepackage{makecell}
\usepackage{microtype}
\definecolor{na}{gray}{0.9}
\usepackage{color}
\usepackage{makecell}
\usepackage{multirow}
\usepackage{xspace}
\usepackage{colortbl}
\usepackage{amsfonts}
\usepackage{inconsolata}
\usepackage{epsfig,graphics,float}
\usepackage{caption}
\usepackage{booktabs}
\usepackage{amsmath}
\usepackage{mathtools}
\usepackage{sidecap}
\usepackage{multirow}
\usepackage{epsfig,graphics,float}
\usepackage{todonotes}
\usepackage[ruled,vlined]{algorithm2e}
\usepackage{subfig,balance,lineno}
\usepackage{float}
\usepackage{verbatim}

\usepackage{wrapfig}
\title{Non-Intrusive Adaptation:\\ Input-Centric Parameter-efficient Fine-Tuning for Versatile Multimodal Modeling}

{\centering
\author{{Yaqing Wang\thanks{Both authors contributed equally} , Jialin Wu${^*}$, Tanmaya Dabral, Jiageng Zhang, Geoff Brown,} \\
{\textbf{Chun-Ta Lu, Frederick Liu, Yi Liang, Bo Pang, Michael Bendersky, Radu Soricut}} \\
{Google}\\
 \texttt{\{yaqingwang, jialinwu, tanmayad, jiageng, geoffbrown, chunta,} \\ \texttt{frederickliu, yiliang, bopang, bemike, rsoricut\}@google.com}}

}

\newcommand{\sysname}{{\tt AdaLink}\xspace}
\iclrfinalcopy % Uncomment for camera-ready version, but NOT for submission.
\begin{document}

\maketitle
\begin{abstract}
Large language models (LLMs) and vision language models (VLMs) demonstrate excellent performance on a wide range of tasks by scaling up parameter counts from O(10$^9$) to O(10$^{12}$) levels and further beyond. These large scales make it impossible to adapt and deploy fully specialized models given a task of interest. Parameter-efficient fine-tuning (PEFT) emerges as a promising direction to tackle the adaptation and serving challenges for such large models. We categorize PEFT techniques into two types: intrusive and non-intrusive. Intrusive PEFT techniques directly change a model's internal architecture. Though more flexible, they introduce significant complexities for training and serving. Non-intrusive PEFT techniques leave the internal architecture unchanged and only adapt model-external parameters, such as embeddings for input. In this work, we describe \sysname as a non-intrusive PEFT technique that achieves competitive performance compared to SoTA intrusive PEFT (LoRA) and full model fine-tuning (FT) on various tasks. We evaluate using both text-only and multimodal tasks, with experiments that account for both parameter-count scaling and training regime (with and without instruction tuning).

\end{abstract}
\section{Introduction}

While large language models (LLMs) \citep{NIPS2017_3f5ee243,T5,brown2020language,chowdhery2022palm,OpenAI2023GPT4TR,anil2023palm}  and vision-language models (VLMs) \citep{alayrac2022flamingo,li2023blip,wang2022image,chen2023pali} have recently demonstrated remarkable capabilities across a variety of tasks, several challenges persist. Due to the prohibitive engineering cost and inefficiencies involved in maintaining separate models for different tasks, it's still an open question how to adapt these models for different specialized use cases to incorporate the latest information. Therefore, there is a trend towards parameter-efficient fine-tuning (PEFT) as a promising solution to these challenges, offering a trade-off between adaptability and efficiency. PEFT techniques, such as adapters~\citep{houlsby, pfeiffer2020AdapterHub, pfeiffer2021adapter}, LoRA~\citep{hu2021lora}, and prompt tuning \citep{prompt_tuning,liu2021p}, introduce only a small percentage of additional parameters for fine-tuning while leaving the bulk of the LLM's parameters unchanged. Within this framework, we differentiate between intrusive and non-intrusive PEFT methods based on the degree to which they interact with or alter the LLM's core architecture, like the transformer blocks.

Intrusive adaptation methods, including LoRA~\citep{hu2021lora}, Adapter~\citep{pfeiffer2021adapter, beck2021adapterhub}, prefix-tuning~\citep{prefix} and their combinational methods \citep{chen2023parameter,mao2021unipelt}, make direct changes  to the model architecture  or the internal parameters flexibly, modifying the existing layers and adding new layers. While offering strong expressive power by flexibility and potentially reducing the performance gap akin to full model fine-tuning, they introduce significant complexities in architecture design spaces and the serving infrastructures. Moreover, these core architectural changes often lead to compatibility issues and complicate the engineering required for the deployment of a single LLM equipped with multiple adaptation components.  
Such intricacies also heighten the possibility of unintended behaviors, for instance, potentially loading incorrect adaptation weights for different tasks or layers, thereby making extensive validation and testing all the more imperative for ensuring model reliability.

%[yaqing's version]In contrast, non-intrusive adaptation strategies including prompt-tuning aim to adjust a model's behavior with minimal changes to its internal architecture or parameters. This is often achieved by modifying the input that is fed into the model or by post-processing the model's outputs.   Additionally,  a non-intrusive adaptation strategy would allow  make granular changes at the input level for each data point in the same batch, enabling the model to remain flexible and adaptable to different customization needs. However, prompt tuning has been shown to struggle with optimization difficulties~\citep{} and is often less effective in adapting models to multi-task scenarios—areas~\citep{} where large language models (LLMs) excel. Toward this challenge, we introduce a novel approach called AdaPrompt, which retains the non-intrusive benefits of prompt tuning by focusing on input-level transformations, avoiding any changes to the internal transformer architecture. Remarkably, we discovered that this simple input-level transformation can achieve performance on par with full model fine-tuning, without encountering the optimization challenges commonly associated with prompt tuning.

In contrast, non-intrusive adaptation strategies like prompt-tuning \citep{prompt_tuning} aim to adjust a model's behavior with minimal changes to the internal architecture or parameters that are often achieved by modifying the input to the core architecture. They typically allow users to make granular changes at the input level for each example in the same batch. As a result, the model remains flexible and adaptable to different customization needs. 
%However, non-intrusive PEFT like propmt-tuning has been shown to struggle with optimization difficulties \citep{razdaibiedina2023residual}, often less effective in adapting models into complicated task setup like multi-task~\citep{wang2022multitask} and is not clear .
However, non-intrusive Parameter Efficient Fine-Tuning (PEFT) methods such as prompt-tuning have encountered optimization challenges~\cite{razdaibiedina2023residual}. They are often less effective in adapting models for complex tasks, such as multi-tasking~\citep{wang2022multitask}, and are still in the exploratory phase for multimodal settings, particularly in preserving the position of vision tokens when processing visual input.
Toward these challenges, we introduce a novel approach called \sysname that introduces an adaptation module situated between the embedding and main transformer blocks of LLMs form as a link, retaining the non-intrusive benefits and alleviating the optimization difficulties.

Recent work~\citep{wei2021finetuned,sanh2021multitask,mishra2022cross,llama2} has demonstrated the ability of large language models (LLMs) to acquire a variety of skills and generalize well to unseen tasks through instruction tuning. In this paper, we explore adapting both raw and instruction-tuned LLMs using parameter-efficient fine-tuning (PEFT). We find that starting from an instruction-tuned checkpoint reduces the amount of adaption parameters needed, facilitating the adaption training process and further improving results. The combination of instruction tuning and PEFT unlocks substantial potential, achieving performance on par with full model fine-tuning on diverse text and multimodal tasks. As instruction-tuned LLMs continue to gain prevalence, non-intrusive PEFT methods like the {\sysname} proposed here suffice to obtain optimized performance and emerge as a practical and effective tuning approach. Empirically, we conducted comprehensive experiments on multi-modal (captioning and VQA) tasks and natural language understanding tasks. By tuning only less than $0.02\%$ of a pre-trained language model’s parameters, {\sysname} reaches competitive or even better results compared to  full model fine-tuning methods.  

\noindent\textbf{Properties of AdaLink.} {\sysname} enables efficient and scalable adaptation through its lightweight yet expressive module design. The added computational complexity grows only linearly with model embedding dimension, invariant to other model parameters. This avoids the quadratic scaling incurred by methods like prompt tuning that increase sequence length. Further, {\sysname} provides flexible partial input adaptation, transforming only selected embeddings to minimize interference across modalities or tasks. The modular nature also affords configurable serving, allowing {\sysname} to act as an intermediate processing unit or directly transform vocabulary embeddings. Overall, {\sysname} delivers customizable and scalable task adaptation while limiting complexity overhead and preserving model architecture, making it highly promising for large-scale deployment.
\section{Background}
\label{sec:background}
\noindent \textbf{Prompt tuning.}
Given a pre-trained language model with parameters $\Theta$ and a target task, full model fine-tuning can be parameter-inefficient for multiple tasks. Prompt tuning (PT) offers a more efficient and non-intrusive alternative by initializing a few learnable prompt vectors and appending them to the input embeddings without touching $\Theta$~\citep{lester2021power} and transformer architecture. This approach optimizes a loss function with respect to the prompt vectors and has shown to be effective. Even though prompt tuning is non-intrusive and easy to deploy, it still suffers from a big performance gap in multi-task settings~\citep{wang2022multitask} and sensitivity to initialization 
%have been observed
~\citep{lester2021power, su2022transferability, zhong2022panda}.

\label{subsec:adapter}

\noindent \textbf{Adapter and LoRA.} Alternatively, adapters~\citep{houlsby} and LoRA~\citep{hu2021lora} can be used  to adapt LLMs for downstream tasks with a small number of additional parameters.
%while maintaining main function unchanged.  
%The adapter and LoRA 
These fine-tuning strategies introduce new parameters into LLMs in an intrusive manner. During fine-tuning,  the new parameters are updated with the original LLM parameters kept frozen. Adapters and LoRA usually consist of two fully connected layers.  As an example, see an illustration of adapter as shown on the right.
%Figure~\ref{fig:houlsby-adapter} for illustration, where 

\begin{wrapfigure}{r}{0.3\textwidth}
%\begin{figure}[h!]
\vspace{-0.1in}
 	\centering
  	\includegraphics[width=1.6in]{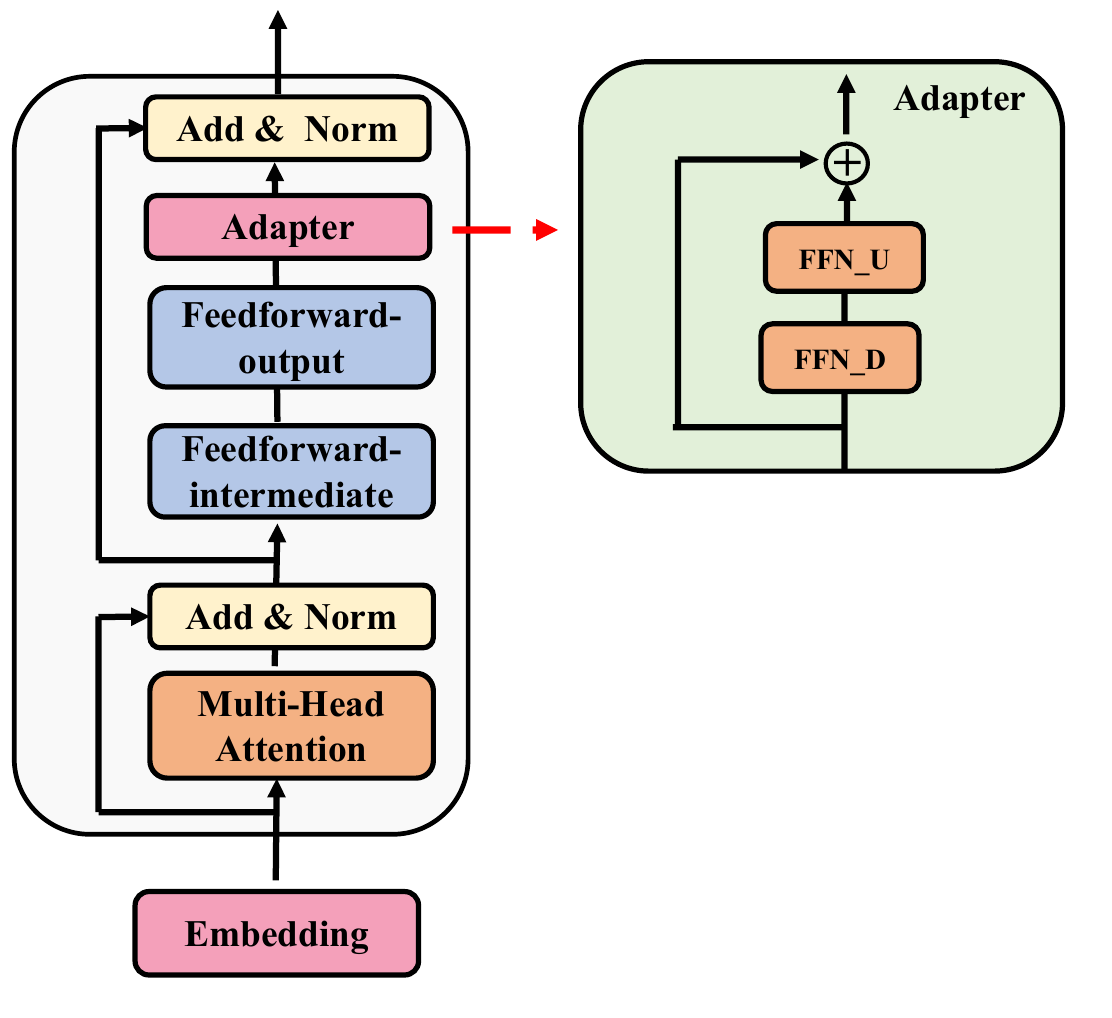}
 	  	%\caption{Conventional adapter design in standard Transformer architecture.}
 	 	 	\label{fig:houlsby-adapter}
 	 	 	
%\end{figure}
\vspace{-0.5in}
\end{wrapfigure}

The adapter layer uses a down projection $\mathcal{W}^{down} \in \mathcal{R}^{d \times r}$ to project input representation $x$ from model dimension $d$ to a low dimensional space $r$ (referred as the bottleneck dimension), followed by a nonlinear activation function $f(\cdot)$, and a up-projection with $\mathcal{W}^{up} \in \mathcal{R}^{r \times d}$ to project the low-dimensional features back to the original dimension.

\section{Methodology}
\subsection{Input Representations}
\noindent {\bf Text Representations.}  For the text representation, we follow the T5~\citep{T5} to use SentencePiece for tokenization, which breaks down the input text into subword units. Let \( T = \{t_1, t_2, ..., t_n\} \) represent the input text, where \( t_i \) is the \( i^{th} \) token and \( n \) is the length of the text.  The tokenized input is passed through an embedding layer to convert into continuous vectors.  Formally, this can be represented as 
$\mathbf{E}_{text} = \{\mathbf{e}_1,  \mathbf{e}_2, ..., \mathbf{e}_n \}$, where $\mathbf{e}_i$ denotes the embedding of token $t_i$.

\noindent {\bf Image Representations.} For the image representations, we follow PaLI \citep{chen2023pali} to use the ViT module to produce visual embeddings. Each image is resized to a fixed size %one resolution according to base model choices and 
and then partitioned into  non-overlapping patches with patch size $14\times14$.
%and each patch can be considered as a separate visual token 
%and fed into the ViT. 
We flatten the output patch-embeddings from the ViT module as the image representations $\mathbf{E}_{image}$.

\noindent {\bf Image-Text Representations.} 
%The image-text representations are to concatenate the image representations $\mathbf{E}_{image}$ and the text representations  $\mathbf{E}_{text}$  as
Visual embeddings and text embeddings are concatenated to form the multimodal input sequence:
$\mathbf{E} =\{\mathbf{E}_{image}, \mathbf{E}_{text}\}$.   \\
\subsection{AdaLink Module}

In essence, \sysname is designed around the concept of incorporating a transformation function as the link between the embedding layer and the main transformer blocks.  This added layer serves as a mechanism for nuanced adaptation. The process begins with data being converted into embeddings through the embedding layer or vision modules. These embeddings are then passed through the  \sysname Modules, resulting in the transformation of the selected inputs. These transformed inputs are subsequently fed into the frozen main transformer blocks for further processing. 
To our surprise, we found that an adapter structure with two fully connected layers is quite effective empirically.
%Empirically, we surprisingly found that utilizing an adapter structure with two fully connected layers is quite effective. 
This approach allows us to achieve competitive results without adding significant complexity, and it maintains several advantageous properties as scalable complexities and versatile deployment strategies that we will discuss in more detail in the subsequent sections. 

More formally, we follow the notation from Sec. \ref{sec:background} to describe {\sysname}, which consists of two fully connected layers. The down projection $\mathcal{W}^{down} \in \mathcal{R}^{d_{emb} \times r}$ projects input representation from the original model dimension $d_{emb}$ to a low dimensional space $r$ (referred to as the bottleneck dimension);
%with $d_{emb}$ being the embedding dimension of the model 
the up-projection with $\mathcal{W}^{up} \in \mathcal{R}^{r \times d_{emb}}$ projects the low-dimensional features back to the original embedding dimension.  {\sysname} has the flexibility to be used as a standalone adaptation module on a per-task basis or on a per-modality basis. We introduce these two scenarios as follows and leave other potential settings  for future research.

\textbf{Multi-task AdaLink.} The conventional parameter-efficient fine-tuning methods were proposed to adapt LLMs to different tasks without creating expensive copies of the original models and storage-efficient. {\sysname} also enables flexibility in the granularity of task adaptation. For example, in multi-task learning scenarios, one can associate a separate {\sysname} module with each task. During training, the input embeddings are selectively transformed by the task-specific {\sysname} before passing through the shared transformer backbone. This targets adaptation to the nuances of each task while enabling positive knowledge transfer through the shared parameters. At inference time, the model routes the inputs through the corresponding task's {\sysname} module to elicit adapted behavior for that task. The rest of the model remains unchanged, avoiding negative interference. Compared to LoRA and Adapter, {\sysname} does not require to architecture modification and further reduce the engineering load extend the functions of LLMs when deploying. Compared to prompt tuning, {\sysname} does not introduce additional cost to the transformer blocks with new tokens.

\textbf{Multimodal AdaLink.} In addition to per-task adaptation, \sysname also enables flexible per-modality adaptation in multimodal settings. For models that take heterogeneous input types like text, image, audio, etc., one can associate a distinct {\sysname} module with each modality. During training and inference, the embeddings for each modality get selectively transformed by their corresponding \sysname before fusion. A key benefit is that this modality-specific adaptation isolates interference across modalities. It also allows the modality representations to be handled independently for greater flexibility; for instance, storing them separately or fusing them at different levels. .
More formally, given an input consisting of an image $\mathbf{x^{image}}$ and text $\mathbf{x^{text}}$, we first obtain modality-specific representations $\mathbf{E}_{image}$ and $\mathbf{E}_{text}$. These are then fed into separate \sysname modules to get adapted embeddings
\begin{align}
\Tilde{\mathbf{E}}_{\text{image}} &= \mathbf{E}_{\text{image}} + f(\mathbf{E}_{\text{image}} \cdot \mathbf{W}^{\text{down}}_{\text{image}}) \cdot \mathbf{W}^{\text{up}}_{\text{image}}, \\
\Tilde{\mathbf{E}}_{\text{text}} &= \mathbf{E}_{\text{text}} + f(\mathbf{E}_{\text{text}} \cdot \mathbf{W}^{\text{down}}_{\text{text}}) \cdot \mathbf{W}^{\text{up}}_{\text{text}},
\end{align}
where $f$ indicates non-linear activation function. We find that removing non-linear activation results in only a negligible decrease in performance, thus we remove it for simplicity. The adapted modality representations $\mathbf{\Tilde{E}}_{image}$ and $\mathbf{\Tilde{E}}_{text}$ are concatenated to form the combined representation $\mathbf{\Tilde{E}}=\{\mathbf{\Tilde{E}}_{image}, \mathbf{\Tilde{E}}_{text}\}$. This $\mathbf{\Tilde{E}}$ is then passed into the main Transformer model for further processing. By transforming each modality separately, {\sysname} provides targeted adaptation while isolating interference across modalities.

\begin{figure}
 	\centering
  	\includegraphics[width=3.2in]{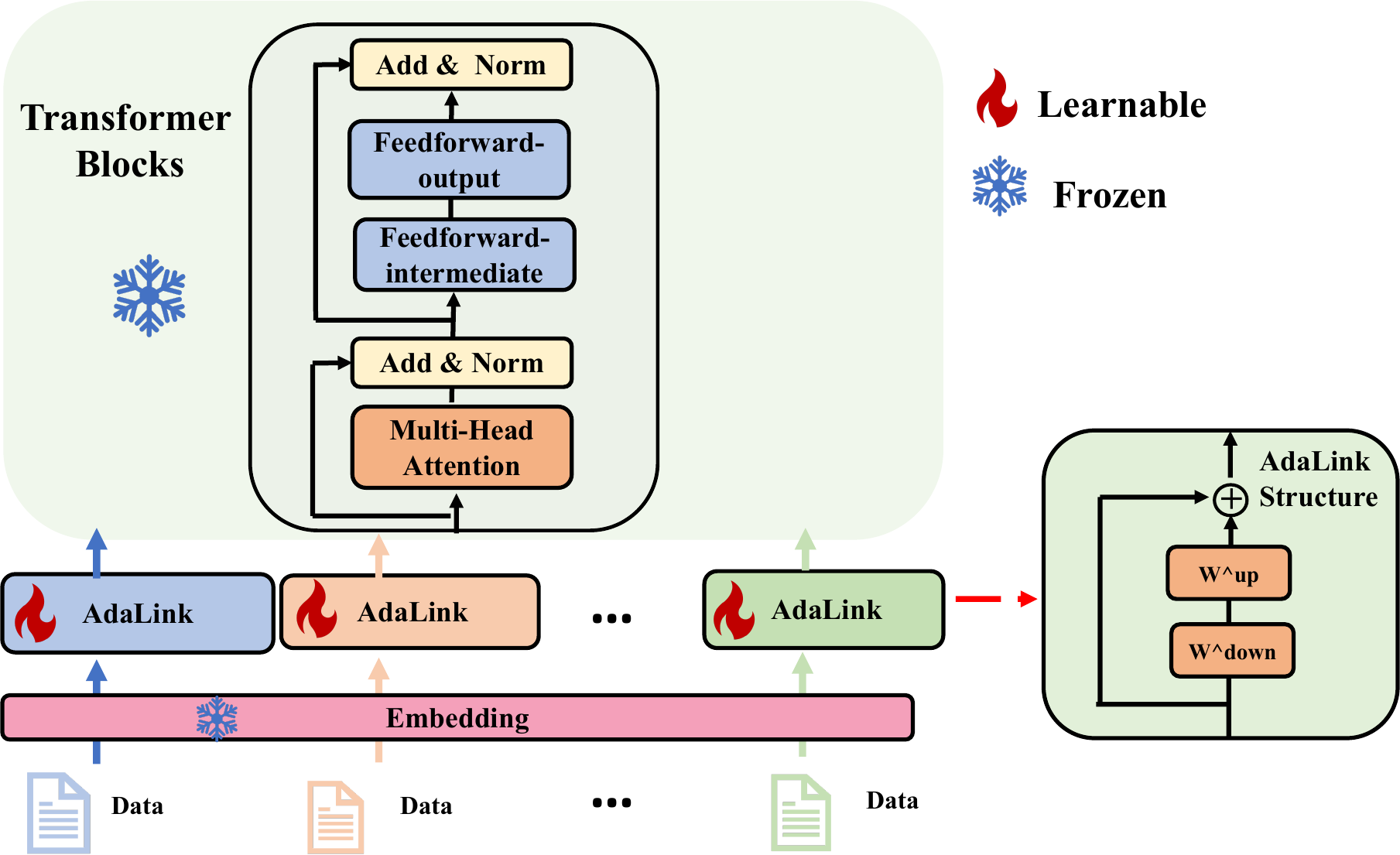}
 	  	\caption{Overview of {\sysname}. Only newly added \sysname~modules are learnable while maintaining other components frozen. The different data is first fed into embedding layer and then goes through the corresponding {\sysname} respectively before main shared Transformer Blocks for adaptation per scenario.  }
 	 	 	\label{fig:adalink}
 	 	 	\vspace{-0.1in}
\end{figure}
% \subsection{Mixture-of-Expert AdaPrompt}

% As discussed above, {\sysname} can be used as an standalone adaptation module per task or per modality. A natural question is if we can build a set of $M$ {\sysname} modules which automatically learn to route to one or more {\sysname} modules for  different tasks. To achieve this design, we consider to bridge {\sysname} and Mixture-of-Expert~\citep{shazeer2017outrageously, fedus2021switch,lepikhin2020gshard,zuo2021taming} and name it as {MoE-\sysname}. In {MoE-\sysname}, {\sysname} is still placed ahead of transformer blocks to maintain non-intrusive benefits instead of following conventional MoE architecture to insert expert parameters into each transformer. We then borrow the routing idea to route different input embeddings to different {\sysname} modules. More specially, we adopt the routing policy design of \citep{fedus2022switch}, which propose several improvements on stabilizing training and reducing communication costs. More formally, the input token embeddings are first fed into a gating network to route each token to a subset of $M$ {\sysname} modules. To note, we still follow the residual design that the outputs of {MoE-\sysname} are as addition to intial input embeddings. While MoE have more {\sysname} modules, the sparsity allows better scaling and better finetuning performance at scale. 

\subsection{Discussion on Properties of {\sysname}}

\textbf{Scalable  Computational Costs.} Consider that we have an input with sequence length of $N$, the embedding dimension of LLMs is $d_{emb}$ and {\sysname} with a rank of $r$, the added complexity is $ \mathcal {O}(Nd_{emb}r)$. The computational complexity of the {\sysname} remains invariant with respect to the scaling of model layers and is linearly proportional to embedding dimension of LLMs. In contrast, prompt tuning appends additional embeddings, thereby increasing the sequence length, which leads to a quadratic increase in computational complexity. This escalation in complexity can be exacerbated with the scaling of large language models (LLMs).

\textbf{Minimal Interference.} A key benefit of {\sysname} is its flexibility in adapting to partial inputs, such as a subset of modalities, without requiring any changes to the main transformer architecture. The adaptation is encapsulated in the lightweight {\sysname} modules that transform selected embeddings before feeding into the standard transformer blocks. Unlike methods that inject additional soft tokens, {\sysname} does not modify the original input representations. This preserves the positional information of inputs like images, where spatial relationships between objects are critical. By limiting adaptation to the {\sysname} modules, {\sysname} allows easily adapting powerful LLMs to new scenarios.

\textbf{Configurable Serving.} {\sysname} can be deployed as an intermediate processing unit as shown in Figure~\ref{fig:adalink}, bringing with it added complexity. Additionally, it can be utilized to transform vocabulary embeddings. In this manner, while the complexity remains constant, there is an associated increase in the storage requirements due to the addition of the embedding layer.

\section{Experiments}

\subsection{Multimodal Experiments}
% checkpoint details: IT
% MMIT
We conduct experiments on four VQA and two image captioning tasks using PaLI-X \citep{chen2023pali}, a 55B multi-modal foundational model that achieved SoTA results on a wide range of vision and language benchmarks. We demonstrate that non-intrusive PEFT achieve very competitive results compared to full model fine-tuning for a large-scale VLM like PaLI-X, especially on a multimodal instruction-tuned variant.

\subsubsection{Base Models}

\textbf{Raw checkpoint:}  We refer to the PaLI-X checkpoint pre-trained per \citep{chen2023pali} with a resolution of 756 $\times$ 756 as the {\em raw} checkpoint. %We also consider a (relatively) small-scale model, a 5B variant of PaLI, pre-trained with similar objectives at a higher resolution (812 $\times$ 812) and better text understanding capabilities.

\textbf{MMIT variant:} We also experiment with a {\em multimodal instruction-tuned (MMIT)} variant, where we finetune the raw PaLI-X checkpoint on MMIT tasks.  
%Due to the limitation and digression of the task scopes in most public academic benchmarks, we create the MMIT task collections 
The MMIT tasks are created in the spirit of ``Self-Instruct'' \citep{wang2022self}, taking advantage of the powerful large language models. We consider three types of tasks: (i) Long-form captioning where multiple captions are generated for each image and LLMs \citep{anil2023palm} are used to combine and summarize them into a longer and more detailed caption; (ii) Creative writing where LLMs are first used to generate novel creative writing prompts and then used to generate actual writings given the prompts based on image captions. (iii) Long-form question answering where LLMs are used to generate questions and answers with rationales given image captions.  Note that these tasks collectively cover a wide variety of usecases rooted in everyday life.  But they are also general in the sense that we do not expect them to be directly in-domain for the downstream tasks considered in this work. In particular, we experiment on down-stream tasks that require specific skills such as understanding scene texts and documents, or answering knowledge intensive questions.

\subsubsection{Implementation Details}
We compare full model fine-tuning (FT) against three types of PEFT: prompt tuning (PT) \citep{prompt_tuning}, LoRA \citep{hu2021lora} and \sysname. We use adafactor \citep{shazeer2018adafactor} as the optimizer. The learning rate is set to 0.03 for PEFT and 0.0001 for fine-tuning with a linear warmup and reciprocal square root decay unless otherwise specified. By default, we set the dropout rate as 0.1 to prevent over-fitting. 

\textbf{Fine-tuning.} Recall that PaLI-X follows the encoder-decoder architecture where image embeddings produced by a ViT module, along with text embeddings, are fed to the multimodal encoder as one sequence.  In full model fine-tuning (FT) experiments, we keep the ViT module frozen and only fine-tune the encoder-decoder backbone.

\textbf{LoRA.} We add LoRA weights on each linear layer in the multi-head attention and the MLP blocks in the encoder transformer blocks for both base models. Similar to \citep{yang2022prompt}, we found that adding LoRA weights in the decoder did not help the adaptation performance much at the cost of twice as many parameters. We use a LoRA rank of 16 in experiments on the raw-checkpoint and a LoRA rank of 4 in experiments on the MMIT variant.

\textbf{Prompt Tuning.}   Prompt Tuning (PT) is implemented by concatenating $64$ soft tunable tokens to the original input sequence, and feeding that concatenated sequence to the multimodal encoder of PaLI-X. We apply two layers of residual re-parameterization \citep{razdaibiedina2023residual} for more stable results. We use a dropout rate of 0.05 for all prompt tuning experiments as we found it to outperform the default rate of 0.1.

\textbf{AdaLink}. We insert modality-specific \sysname modules to the embeddings of the text tokens and the visual tokens as a non-intrusive PEFT technique for the base model. We use a rank of $64$ in all the experiments.

\subsubsection{Image captioning Results}

% \begin{table}[h]
% \centering
% \begin{tabular}{l|cc|cc|cc}
% \toprule
% &  \multicolumn{2}{c|}{COCO} & \multicolumn{2}{c|}{Textcaps} & \multicolumn{2}{c}{avg. $\delta$ to FT} \\ \midrule
%                 & RAW         & MMIT       & RAW           & MMIT         & raw        & MMIT   \\\hline
% FT              & 147.4       & 147        & 148.6         & 148.5        & 0          & 0        \\ 
% LoRA            & 146.1       & 146.8      & 147.8         & 148.6        & -1.05      & -0.05    \\ \hline
% Prompt-tuning   & 143.5       & 142.2      & 144.9         & 145.5        & -3.8       & -3.9     \\
% \sysname        & 146.2       & 146.3      & 145.2         & 147.9        & -2.3       & -0.65    \\
% \bottomrule
% \end{tabular}
% \caption{PEFT results on COCO captioning karpathy test set and textcaps captioning validation set. We report cider score for each task. \sysname consistently outperforms prompt tuning as non-intrusive PEFT approaches and achieves competitive results to FT.}
% \label{tab:mm_cap_results}
% \end{table}

\begin{table}[h]
\vspace{-0.05in}
\centering
\small
\caption{PEFT results on COCO captioning Karpathy test set and TextCaps captioning validation set. We report cider score for each task. \sysname  consistently outperforms the other non-intrusive PEFT approach (prompt tuning)  and achieves competitive results to fine-tuning.  $^\dag$Recall we keep the ViT module frozen; 32B is the parameter count for the encoder-decoder backbone.}
	\resizebox{0.9\linewidth}{!}{
\begin{tabular}{l|c|c|cc|cc|cc}
\toprule
&  Non-intru- & \# params & \multicolumn{2}{c|}{COCO} & \multicolumn{2}{c|}{TextCaps} & \multicolumn{2}{c}{avg. $\delta$ to FT} \\ 
               & sive & & MMIT        & RAW        & MMIT          & RAW          & MMIT       & raw   \\\midrule
Fine-tuning (FT)  & No &    32B$^\dag$   & 147         & 147.4      & 148.5         & 148.6        & 0          & 0        \\ 
LoRA        & No &    19M       & 146.8       & 146.1      & 148.6         & 147.8        & -0.05      & -1.05    \\ \hline
Prompt-tuning (PT)  & Yes &   262k  & 142.2       & 143.5      & 145.5         & 144.9        & -3.9       & -3.8     \\
\sysname     & Yes &    1.05M      & 146.3       & 146.2      & 147.9         & 145.2        & -0.65      & -2.3     \\
\bottomrule
\end{tabular}}
\label{tab:mm_cap_results}
\end{table}

Table \ref{tab:mm_cap_results} reports PEFT image captioning CIDEr scores \citep{vedantam2015cider} on COCO \citep{lin2014microsoft} and TextCaps \citep{sidorov2020textcaps}. Within the non-intrusive PEFT family, \sysname outperforms prompt tuning by about 2 cider points on average, indicating the effectiveness of directly adapting the input embeddings. 

More importantly, we observe smaller gaps between \sysname and FT on the MMIT variant than the raw checkpoint.  This is consistent with our hypothesis that \sysname can benefit more from instruction tuned base models, enabling competitive results to FT (an average of difference of 0.65).  It is impressive for \sysname (1.05M parameters to tune) to come within one point of full fine-tuning (32B parameters to tune).  Indeed, given the much smaller number of tunable parameters, non-intrusive PEFT may suffer from less expressive power.  This is perhaps less of a problem given the expressive power in large-scale base models (like PaLI-X) themselves, and partly further mitigated when base models are pre-trained on a larger variety of tasks (e.g., the MMIT variant in our experiments).  Note also: while PaLI-X provides a very strong base model, with SoTA finetuning results on a wide array of benchmarks, it's not strong to the point where this level of performance can easily be achieved with zero tuning.  As a reference point, on the same COCO Captions task, \citet{chen2023pali} reported a CIDEr score of 107.6 for 4-shots and 114.5 for 32-shots learning, a difference of more than 30 points to FT.  Thus reaching SoTA FT performance with light-weight tuning technique like \sysname is non-trivial.

While LoRA also gets better performance over the MMIT variant, the performance gap between \sysname and LoRA is also smaller on this variant.  Given the increasing popularity of instruction tuned LLMs, non-intrusive PEFT, especially \sysname, become a strong candidate with significantly lower complexities in architecture and serving infrastructure at the cost of very minor performance degradation.  As multimodal instruction tuning tasks become more comprehensive and diverse, we hypothesize there can be even smaller performance gaps between simple non-intrusive PEFT approach like \sysname and intrusive PEFT or full model fine-tuning.  In the case of increased base model size, complexities of non-intrusive PEFT approaches like \sysname do not grow with the depth of the growing models, presenting another clear advantage in terms of practicality.

Next we present additional ablation studies on COCO Captions, again reporting results on the Karparthy test split.

\begin{table}[!hbt]
\centering
\caption{Effect of rank in \sysname on the COCO captioning task. }
\begin{tabular}{l|cccc}
\toprule
Rank        &  4  & 16   & 64   & 256   \\ \hline
CIDEr &  144.5 & 145.3   & 146.3   & 146.3   \\ \bottomrule  
\end{tabular}
\label{tab:rank_COCO}
\end{table}

\textbf{Effect of the rank.} Table \ref{tab:rank_COCO} reports the effects of changing the ranks in \sysname using the MMIT variant. We observe that the performance is not very sensitive to  rank, indicating the stability of  \sysname. Even a rank of 4 can help the models adapt to reasonable performance, and the performance saturated at a rank of 64.

\begin{table}[!hbt]
\centering
\caption{Effect of separately adapting the input embeddings in each modality}
	\resizebox{0.6\linewidth}{!}{
\begin{tabular}{l|cc|cc}
\toprule
        & \multicolumn{2}{c|}{Single unified \sysname}  & \multicolumn{2}{c}{Modality-based \sysname} \\ \hline
        & MMIT   & Raw         & MMIT  & Raw \\ \hline
CIDEr & 145.5  & 145.2       & 146.3  & 146.2    \\ \bottomrule  
\end{tabular}}
\label{tab:modality_ablatation}
\vspace{-0.1in}
\end{table}

\textbf{Effect of using separate adapters for image and text modalities}
Next, we compare the default modality-based \sysname with separate adapters for image and text modalities to a baseline that uses one unified \sysname adapter with rank 128 (twice as much as the default \sysname) to adapt both visual and text tokens.
%We examine the  modality-based \sysname 
%help adaptation in COCO captioning using PaLI-X in 
Table \ref{tab:modality_ablatation} presents their performance on COCO captioning.  Regardless of the base model variant used, modality-based \sysname outperforms the single unified \sysname by about 1 CIDEr point while using the same number of additional parameters, quantifying the benefit of modality-specific modeling, something prompt tuning struggles to achieve.

% \begin{table}[h]
% \centering
% \begin{tabular}{l|cc|cc}
% \toprule
%         & \multicolumn{2}{c|}{Single unified \sysname}  & \multicolumn{2}{c}{Modality-based \sysname} \\ \hline
%         & Raw    & MMIT        & Raw  & MMIT \\ \hline
%  PaLI-X &  145.2 & 145.5      & 146.2  & 146.3    \\ \bottomrule  
% \end{tabular}
% \caption{Effect of separately adapting the input embeddings in each modality}
% \label{tab:modality_ablatation}
% \end{table}

\subsubsection{VQA Results}
In Table \ref{tab:mm_vqa_results}, we present VQA performance using PEFT on four VQA tasks: OK-VQA \citep{marino2019ok} which requires drawing upon outside knowledge, DocVQA \citep{mathew2021docvqa} which examines document understanding capabilities, and two scene-text understanding datasets --- TextVQA \citep{singh2019towards} and ST-VQA \citep{biten2019scene}. We follow standard evaluation metrics, using soft accuracy \citep{antol2015vqa} for OKVQA and TextVQA and ANLS score for DocVQA and ST-VQA.

% \begin{table}[h]
% \scriptsize
% 	\resizebox{1.0\linewidth}{!}{
% \begin{tabular}{lc|c|cccccccccc}
% \toprule
% Tuning  &Non-intru- & \# params & \multicolumn{2}{c}{OKVQA} & \multicolumn{2}{c}{DocVQA} & \multicolumn{2}{c}{ST-VQA} & \multicolumn{2}{c}{TextVQA} & \multicolumn{2}{c}{avg. $\delta$ to FT}  \\
% Approach   & sive &          & Raw        & MMIT         & Raw        & MMIT          & Raw          & MMIT        & Raw        & MMiT           & RAW         & MMIT            \\\midrule
% FT  & No& 32B$^\dag$       & 66.1        & 66.9        & 80.0         & 82.8        & 80.2         & 79.7        & 71.9         & 70.7         & 0.0           & 0.0       \\     
% LoRA        & No& 19M    & 63.3        & 67.1        & 80.6         & 83.2        & 78.6         & 80.0        & 69.1         & 70.8         & -1.7          & +0.25         \\
% Prompt tuning  &Yes& 262k     & 64.9        & 66.4        & 79.7         & 82.4        & 78.3         & 79.8        & 69.7         & 70.4         & -1.4          & -0.3           \\
% \sysname     &Yes & 1.05M    & 63.9        & 66.8        &    78.3     & 82.9        & 77.9         & 80.0        &      67.8        & 70.2         &     -2.58          & -0.05          \\ \bottomrule

% \end{tabular}}
% \caption{PEFT results on four VQA tasks on the validation splits using the PaLI-X models. $^\dag$ We only fine tune the Encoder-Decoder part of the PaLI model and leave the ViT part frozen. }
% \label{tab:mm_vqa_results}
% \end{table}

\begin{table}[h]
\scriptsize
\caption{PEFT results on four VQA tasks on the validation splits.}
	\resizebox{1.0\linewidth}{!}{
\begin{tabular}{l|c|cccccccccc}
\toprule
  &   \# params & \multicolumn{2}{c}{OKVQA} & \multicolumn{2}{c}{DocVQA} & \multicolumn{2}{c}{ST-VQA} & \multicolumn{2}{c}{TextVQA} & \multicolumn{2}{c}{avg. $\delta$ to FT}  \\
   &           & MMIT        & Raw         & MMIT        & Raw          & MMIT        & Raw          & MMIT        & Raw           & MMIT         & RAW            \\\midrule
FT  &  32B       & 66.9        & 66.1        & 82.8         & 80.0        & 79.7         & 80.2        & 70.7         & 71.9         & 0.0           & 0.0       \\     
LoRA        &  19M    & 67.1        & 63.3        & 83.2         & 80.6        & 80.0         & 78.6        & 70.8         & 69.1         & +0.25          & -1.7         \\
PT  & 262k     & 66.4        & 64.9        & 82.4         & 79.7        & 79.8         & 78.3        & 70.4         & 69.7         & -0.3          & -1.4           \\
\sysname     & 1.05M    & 66.8        & 63.9        & 82.9         & 78.3        & 80.0         & 77.9        & 70.2         & 67.8         & -0.05          & -2.58          \\ \bottomrule

\end{tabular}}
\label{tab:mm_vqa_results}
\end{table}

%For the PaLI-5B raw model, LoRA as an intrusive PEFT approach achieves the most competitive average results to the FT results, yielding only 1.17 performance degradation where both prompt tuning and \sysname achieve worse results, yielding 1.74 and 2.93 performance degradation. We attribute this to the strong expression capabilities offered by the flexibility in modifying the internal architecture. Within non-intrusive PEFT, \sysname falls behind prompt tuning by 1.2 points. We hypothesize that the adapters in \sysname from different modalities learn to separately over fit modality bias rather than collaborating to learn to reason the visual questions given the image. 

As shown in Table \ref{tab:mm_vqa_results}, tuning the MMIT variant in general leads to better performance than tuning the raw checkpoint.  In fact, when using the MMIT variant, the average performance differences among different tuning techniques are negligible, and
\sysname again emerges as an excellent choice due to its ease of serving and lower parameter counts, trailing FT by only 0.05, echoing what we saw from the captioning experiments.
%and \sysname is only 0.05 below FT, %highlighting that \sysname emerges 
%again shown to be an excellent choice considering the ease of serving and the %parameter counts. 

It is worth noting that all three PEFT approaches, both intrusive and non-intrusive, achieved better performance on the MMIT variant, making them competitive with FT.  
%This reveals and supports the trend towards PEFT where the LLMs and VLMs are increasingly powerful and only little adaptation is needed to achieve optimized performance for specialized use cases, relieving the need of directly learning the domain knowledge. 
This again points to an interesting emerging trend:
the increasing power of LLMs and VLMs allows lightweight PEFT adaptation to achieve competitive performance for highly specialized use cases;
moreover, this also enables non-intrusive PEFT approaches like \sysname to perform competitively against intrusive ones.
%Comparing to intrusive PEFT that are more similar to FT in spirits of adapting internal parameters, non-intrusive PEFT approaches like \sysname are often sufficient and offer similar performances with significant ease of model deployment and constant parameter counts with respect to the model depth. 

% \input{subfiles/_old_pali_results}

% 1. Main performance table
% @yaqingwang
% Raw T5: GLUE Tydiaqa QA
% FLAN 

% QA: HotpotQA (Yang et al., 2018), NewsQA (Trischler
% et al., 2017) and SearchQA (Dunn et al., 2017) from MRQA (Fisch et al., 2019); WinoGrande (Sakaguchi et al., 2021), Yelp-2 (Zhang et al., 2015), SciTail (Khot et al., 2018) and PAWS-Wiki

% @jialinwu
% Raw
% MMIT
% Dataset: COCO, textcaps, VATEX, VQA-v2, OKVQA, ST-VQA, NextQA

% 1. Good performance
% 2. IT reduce parameter needed

% 2. Baseline
% lora, prompt tuning, adaprompt 

% 1. Text: T5 xxl + LoRA ? @yaqingwang
% 2. Multimodal LoRA: READY 

% lora, prompt tuning--GLUE

% 3. Difficult to optimize @yaqingwang
% Multi-task for prompt tuning, adaprompt
% Small checkpoint with T5 large to show better peformance
% https://openreview.net/pdf?id=Nk2pDtuhTq

% 4. Serving cost/inference speed:
% 1. LoRA v.s. Adaprompt 
% @jialin

% 5. Few-shot (16 shots, 32 shots, 128 shots) in IT models
% i. multimodal @jialin
% ii. text @yaqingwang

% 6. adaprompt multimodal albation 

% 7 server inference speed 

\subsection{Natural Language Experiments}

\noindent\textbf{Experimental setting.} We perform experiments on a wide range of tasks including eight natural language understanding (NLU) tasks in the General Language Understanding Evaluation (GLUE) benchmark \citep{wang2019glue}. We compare
{\sysname} to full model fine-tuning with various checkpoints including instruction-tuned checkpoint FLAN~\citep{wei2021finetuned} and T5 checkpoints with additional adaption steps following ~\citep{prompt_tuning}. Unless otherwise specified, all of the experiments in this work utilize the 11 billion parameter T5 or FLAN checkpoint as the base model.

\noindent{\bf AdaLink implementation details.} We implement {\sysname} in Jax for experiments. {\sysname} uses a dimension $r$ of $4$ and $256$ with FLAN and T5 checkpoint in single task setting. In multi-task setting, we increase the dimensions to $256$ and $1024$ for FLAN and T5 checkpoints respectively. We found that most of tasks are not sensitive to rank of {\sysname} and  the performance of {\sysname} plateaus after the modules reach a certain size. Increasing the capacity beyond this point yields diminishing returns, with little to no improvement observed in the end task metrics. The learning rate is set to 0.001 for {\sysname}. By default, we set the dropout rate as 0.1 to prevent over-fitting.

\begin{table*}[!t]
%\vspace{-0.1in}
	\caption{Results for NLU tasks on GLUE development set with 11B T5 and FLAN checkpoints. The best result on each task is in \textbf{bold}. Pearson refers to Pearson correlation. 
	\#Param. denotes the number of tunable adaptation parameters. FT indicates full model fine-tuning, which is usually regarded as a upper bound performance for adaptation scenarios.}

	\begin{center}
	\resizebox{1.0\linewidth}{!}{
		\begin{tabular}{l|c|lc c cccccc  c c}
			\toprule \bf Setting & Checkpoint & Method& \#Tunable Param.&MNLI        &QNLI               &SST2          &QQP          &MRPC               &CoLA           &RTE        &STS-B&\bf Avg. \\ 
			                           &&&&Acc             &Acc             &Acc          &Acc            &Acc             &Mcc            &Acc            &Pearson    \\ \midrule

		    \multirow{4}{*}{\bf Single Task} & \multirow{2}{*}{FLAN} & FT &11B x 8 & \textbf{92.1} & 96.0 & 97.1 & 92.2 & 92.2 & 70.1 & 93.9 & 91.2 & 90.6\\
		    &&{\sysname} & 0.008M x 8 & 91.7 & 96.1 & 97.4&90.7& 91.9& 70.0 & \textbf{94.9} & \textbf{93.0} & \textbf{90.7}\\
		    \cline{2-13} 
		   &\multirow{2}{*}{T5}&FT &11B x 8 & 91.8 & \textbf{96.2} & 97.3 & 92.2 & 90.9 & \textbf{72.2} & 92.1 & 91.4  & 90.5\\
		    &&{\sysname} & 0.5M x 8  &  91.4 & 96.0 & 97.1 & \textbf{92.3} & 91.5 & 64.8 & 93.5 & 91.4 & 89.8\\
		    \midrule
		       \multirow{4}{*}{ \bf Multi-Task} & \multirow{2}{*}{FLAN} & FT &11B & 91.2&96.1&97.1 & 91.9 & 90.2 & 70.2 & 93.5 & 89.5& 90.0\\
		    &&{\sysname} &0.5M &91.8 & 95.6 & 96.8 & 90.8 & \textbf{93.1} & 64.5 & 93.1 & 92.7 & 89.8 \\
		   %  &&{MOE-\sysname} &2M & \\
		     \cline{2-13} 
		   &\multirow{2}{*}{T5}&FT &11B & 91.7 & 96.1 & \textbf{97.5} & 90.8 & 90.0 & 65.8 & 89.9 & 87.6 & 88.7  \\
		    &&{\sysname} &2M & 90.1 & 93.8 & 96.0 & 91.2 & 88.0 & 60.0 & 86.3 & 89.9 & 86.9\\
		   % &&{MOE-\sysname} &8M &  91.6 & 96.0 & 96.7 & 91.7 & 89.0 & 69.7 & 93.9 & 91.5 & 90.0\\

			\bottomrule

		\end{tabular}}
	\end{center}

		\vspace{-0.in}
\end{table*}

\noindent\textbf{Single task.} The table compares full fine-tuning versus using {\sysname} for adapting 11B T5 and FLAN checkpoints to individual GLUE tasks. For full fine-tuning, all 11 billion parameters are tuned on each task. With {\sysname}, only the small adapter modules with 0.5-0.008 million parameters are tuned per task. We observe that {\sysname} achieves comparable or better performance than full fine-tuning on most tasks, despite tuning far fewer parameters. For example, with the FLAN checkpoint, {\sysname} attains higher accuracy on SST-2, QQP, RTE and STS-B benchmarks. Overall, {\sysname} achieves a similar average GLUE score to full fine-tuning of 90.7 using FLAN, while only tuning 0.008M adaption parameters per task. This demonstrates {\sysname}'s effectiveness in targeted task adaptation for large language models. The results validate {\sysname} as an efficient and performant approach to adapting pretrained models to individual tasks, without compromising on model capacity. The modular architecture allows for the extension of new tasks or knowledge without the need to redevelop the main models, akin to adding patches to software during version changes.

\noindent\textbf{Multi-task.} Prior work has shown that prompt tuning approaches have optimization difficulties when applied to multiple tasks simultaneously~\citep{wang2022multitask}. As an input-centric method similar to prompt tuning , exploring  capabilities and limits of {\sysname} in the multi-task setting is informative and can help unveil the potential of this new method. {\sysname} exhibits a minor gap of only 1-2\% versus full fine-tuning and it achieves comparable or higher accuracy than full tuning on 6 out of 8 GLUE tasks using the FLAN checkpoint. The gap is most noticeable on the challenging CoLA task requiring complex linguistic adaptations. However, {\sysname}'s strong performance on most benchmarks shows that input-level tuning can effectively emulate task-specific behaviors.

\begin{table*}[!htb]
\vspace{-0.0in}
	\caption{Results for NLU tasks on GLUE development set with 11B T5 and FLAN checkpoints. The performance is reported with respect to varying rank dimensions of {\sysname}.}

	\begin{center}
	\resizebox{0.75\linewidth}{!}{
		\begin{tabular}{l|c|lc c cccccc  c c}
			\toprule \bf Checkpoint & Rank $r$ &MNLI        &QNLI               &SST2          &QQP          &MRPC               &CoLA           &RTE        &STS-B&\bf Avg. \\ 
			                           &&Acc             &Acc             &Acc          &Acc            &Acc             &Mcc            &Acc            &Pearson    \\ \midrule

		    \multirow{3}{*}{\bf FLAN} &  2    & 91.9 & 96.1 & 97.1 & 91.0 & 91.4 & 68.7 & 94.9 & 92.8 & 90.5 \\
        &4    & 91.7 & 96.1 & 97.4 & 90.7 & 91.9 & 70.0 & 94.9 & 93.0 & 90.7 \\
        &8    & 92.0 & 96.2 & 97.3 & 90.8 & 92.2 & 68.9 & 94.6 & 93.0 & 90.6 \\ 
		    \midrule
		       \multirow{4}{*}{ \bf T5} &   64   & 91.5 & 95.9 & 97.3 & 91.7 & 90.2 & 63.3 & 93.1 & 91.8 & 89.3 \\
        &256  & 91.4 & 96.0 & 97.1 & 92.3 & 90.7 & 64.8 & 93.5 & 90.6 & 89.7 \\
        &512  & 91.3 & 96.0 & 97.4 & 92.2 & 91.5 & 62.8 & 93.5 & 91.4 & 89.6 \\
        &1024 & 91.4 & 96.0 & 97.1 & 91.5 & 91.6 & 63.1 & 93.1 & 91.9 & 89.6 \\

			\bottomrule

		\end{tabular}}
	\end{center}

	\vspace{-0.in}
\end{table*}

\noindent\textbf{Analysis of rank.} Our experiments demonstrate that {\sysname} is not very sensitive to the rank hyperparameter. With an instruction-tuned FLAN checkpoint, a small rank of 4 achieves maximum GLUE performance, indicating compact {\sysname} suffice for embedding space transformation. Increasing rank further shows negligible gains, underscoring the stability of {\sysname} architecture. A larger rank is needed for the non-specialized T5 checkpoint, but performance stabilizes quickly. Overall, {\sysname} attains strong adaptation with minimal parametrization across diverse initializations.

\section{Related Work}
The wide scope of capabilities achieved by LLMs \citep{T5,brown2020language,chowdhery2022palm,anil2023palm} and VLMs \citep{alayrac2022flamingo,li2023blip,wang2022image,chen2023pali} comes along with the scaling up of the parameter counts to billion level. This prohibits the conventional model deployment pipelines where different tasks own different copies of the entire model that are served separately. We briefly introduce two means in the following sections for tackling this problem.

\subsection{Instruction Tuning}
Instruction tuning \citep{wei2021finetuned,chung2022scaling,sanh2021multitask,wang2022self,ouyang2022training,longpre2023flan} aims at solving a wide range of tasks using one foundation model. The entire model is fine-tuned on a large mixture of instructions formulated from the tasks of interest. \citep{wei2021finetuned} explore combining 62 NLP datasets with 10 instructions for each set as training data. \cite{chung2022scaling} further expands the scope up to 1800 tasks. The LLMs demonstrate strong capabilities in learning to interpolate the tasks used and generalize well to unseen tasks. As the instruction tuning data size is often limited, recent research proposes ``Self-Instruct'' \citep{wang2022self} that collects data by bootstrapping off their own generations, relieving the annotation burden.

\textbf{Multi-Modal Instruction Tuning}.
Similar to text-only instruction tuning, Multi-Modal Instruction Tuning (MMIT) aims to jointly learn a large collection of visual language tasks. However, as most available vision-language tasks are short captioning, question-answering, and grounding for academic benchmarks that are limited in both visual scope (i.e. covered visual domains) and task scopes. Lots of tasks digress the natural use cases such as storytelling, descriptive caption generation, answering questions with explanations, etc. Therefore, most MMIT \citep{liu2023visual,zhang2023llama,instructblip,gao2023llama} relies on ``Self-Instruct'' \citep{wang2022self} protocols that create training tasks automatically. 

\subsection{Parameter Efficient Fine Tuning}
Instead of deploying specialized full models, recent research investigates more on the parameter-efficient fine-tuning (PEFT) that only adapts a tiny portion of parameters, keeping most of the parameters frozen. We categorized the PEFT approaches into intrusive and non-intrusive approaches. 

\textbf{Intrusive PEFT} makes direct changes to the model architectures, usually to the transformer blocks. Layer-wise prompt tuning \citep{liu2021p} and LLaMA \citep{zhang2023llama} prepend tunable tokens to the transformer blocks' inputs. Adapters \citep{houlsby, pfeiffer2020AdapterHub, pfeiffer2021adapter} insert low-rank MLPs in each block. LoRA \citep{hu2021lora} takes a step further and adds low-rank weights in each linear layer within the self-attention and the MLPs. Though the intrusive PEFT approaches offer more flexibility in design, they introduce significant challenges in model deployment where the adaptation weights need to be transferred to the internal architecture. Besides, the size of the tunable parameters still grows proportionally to the model size. 

\textbf{Non-intrusive PEFT} is input-centric which keeps the core transformer blocks frozen, including both the pre-trained parameters and the computation graph. Prompt tuning \cite{} is the classic example where the tunable tokens are prepended to the word embeddings before being fed into the transformer blocks. However, Experiments show that prompt tuning struggles with optimization difficulties \cite{razdaibiedina2023residual}, requiring a large number of training examples. We propose  \sysname that adapts the input embeddings using low-rank MLPs, taking the benefit of ``zero init'' that avoids the disturbance at the beginning of training. We show that the \sysname achieves competitive results as the full-model fine-tuning with scaling up of the model size.

\section{Conclusions}
In this paper, we examine the influence of scaling up both model parameter counts and pre-training tasks to parameter efficient tuning (PEFT) on both text only and multimodal down-stream tasks. We show that the performance gap between full model fine tuning and PEFT are significantly narrowed with the help of both. This indicates the increasingly powerful LLMs and VLMs only require a slight adaptation, and input-centric  non-intrusive PEFT is often enough to obtain optimized performance and enjoys the ease of deployment and constant size with respect to model depth. We also introduce \sysname that achieves better adaptation performance than prompt tuning within the non-intrusive PEFT family.

\bibliography{iclr2024_conference}

\begin{thebibliography}{48}
\providecommand{\natexlab}[1]{#1}
\providecommand{\url}[1]{\texttt{#1}}
\expandafter\ifx\csname urlstyle\endcsname\relax
  \providecommand{\doi}[1]{doi: #1}\else
  \providecommand{\doi}{doi: \begingroup \urlstyle{rm}\Url}\fi

\bibitem[Alayrac et~al.(2022)Alayrac, Donahue, Luc, Miech, Barr, Hasson, Lenc,
  Mensch, Millican, Reynolds, et~al.]{alayrac2022flamingo}
Jean-Baptiste Alayrac, Jeff Donahue, Pauline Luc, Antoine Miech, Iain Barr,
  Yana Hasson, Karel Lenc, Arthur Mensch, Katherine Millican, Malcolm Reynolds,
  et~al.
\newblock Flamingo: a visual language model for few-shot learning.
\newblock \emph{Advances in Neural Information Processing Systems},
  35:\penalty0 23716--23736, 2022.

\bibitem[Anil et~al.(2023)Anil, Dai, Firat, Johnson, Lepikhin, Passos, Shakeri,
  Taropa, Bailey, Chen, et~al.]{anil2023palm}
Rohan Anil, Andrew~M Dai, Orhan Firat, Melvin Johnson, Dmitry Lepikhin,
  Alexandre Passos, Siamak Shakeri, Emanuel Taropa, Paige Bailey, Zhifeng Chen,
  et~al.
\newblock Palm 2 technical report.
\newblock \emph{arXiv preprint arXiv:2305.10403}, 2023.

\bibitem[Antol et~al.(2015)Antol, Agrawal, Lu, Mitchell, Batra, Zitnick, and
  Parikh]{antol2015vqa}
Stanislaw Antol, Aishwarya Agrawal, Jiasen Lu, Margaret Mitchell, Dhruv Batra,
  C~Lawrence Zitnick, and Devi Parikh.
\newblock Vqa: Visual question answering.
\newblock In \emph{Proceedings of the IEEE international conference on computer
  vision}, pp.\  2425--2433, 2015.

\bibitem[Beck et~al.(2021)Beck, Bohlender, Viehmann, Hane, Adamson, Khuri,
  Brossmann, Pfeiffer, and Gurevych]{beck2021adapterhub}
Tilman Beck, Bela Bohlender, Christina Viehmann, Vincent Hane, Yanik Adamson,
  Jaber Khuri, Jonas Brossmann, Jonas Pfeiffer, and Iryna Gurevych.
\newblock Adapterhub playground: Simple and flexible few-shot learning with
  adapters.
\newblock \emph{arXiv preprint arXiv:2108.08103}, 2021.

\bibitem[Biten et~al.(2019)Biten, Tito, Mafla, Gomez, Rusinol, Valveny,
  Jawahar, and Karatzas]{biten2019scene}
Ali~Furkan Biten, Ruben Tito, Andres Mafla, Lluis Gomez, Mar{\c{c}}al Rusinol,
  Ernest Valveny, CV~Jawahar, and Dimosthenis Karatzas.
\newblock Scene text visual question answering.
\newblock In \emph{Proceedings of the IEEE/CVF international conference on
  computer vision}, pp.\  4291--4301, 2019.

\bibitem[Brown et~al.(2020)Brown, Mann, Ryder, Subbiah, Kaplan, Dhariwal,
  Neelakantan, Shyam, Sastry, Askell, Agarwal, Herbert-Voss, Krueger, Henighan,
  Child, Ramesh, Ziegler, Wu, Winter, Hesse, Chen, Sigler, Litwin, Gray, Chess,
  Clark, Berner, McCandlish, Radford, Sutskever, and Amodei]{brown2020language}
Tom Brown, Benjamin Mann, Nick Ryder, Melanie Subbiah, Jared~D Kaplan, Prafulla
  Dhariwal, Arvind Neelakantan, Pranav Shyam, Girish Sastry, Amanda Askell,
  Sandhini Agarwal, Ariel Herbert-Voss, Gretchen Krueger, Tom Henighan, Rewon
  Child, Aditya Ramesh, Daniel Ziegler, Jeffrey Wu, Clemens Winter, Chris
  Hesse, Mark Chen, Eric Sigler, Mateusz Litwin, Scott Gray, Benjamin Chess,
  Jack Clark, Christopher Berner, Sam McCandlish, Alec Radford, Ilya Sutskever,
  and Dario Amodei.
\newblock Language models are few-shot learners.
\newblock In H.~Larochelle, M.~Ranzato, R.~Hadsell, M.F. Balcan, and H.~Lin
  (eds.), \emph{Advances in Neural Information Processing Systems}, volume~33,
  pp.\  1877--1901. Curran Associates, Inc., 2020.
\newblock URL
  \url{https://proceedings.neurips.cc/paper/2020/file/1457c0d6bfcb4967418bfb8ac142f64a-Paper.pdf}.

\bibitem[Chen et~al.(2023{\natexlab{a}})Chen, Zhang, Shi, Li, Smola, and
  Yang]{chen2023parameter}
Jiaao Chen, Aston Zhang, Xingjian Shi, Mu~Li, Alex Smola, and Diyi Yang.
\newblock Parameter-efficient fine-tuning design spaces.
\newblock \emph{arXiv preprint arXiv:2301.01821}, 2023{\natexlab{a}}.

\bibitem[Chen et~al.(2023{\natexlab{b}})Chen, Djolonga, Padlewski, Mustafa,
  Changpinyo, Wu, Ruiz, Goodman, Wang, Tay, et~al.]{chen2023pali}
Xi~Chen, Josip Djolonga, Piotr Padlewski, Basil Mustafa, Soravit Changpinyo,
  Jialin Wu, Carlos~Riquelme Ruiz, Sebastian Goodman, Xiao Wang, Yi~Tay, et~al.
\newblock Pali-x: On scaling up a multilingual vision and language model.
\newblock \emph{arXiv preprint arXiv:2305.18565}, 2023{\natexlab{b}}.

\bibitem[Chowdhery et~al.(2022)Chowdhery, Narang, Devlin, Bosma, Mishra,
  Roberts, Barham, Chung, Sutton, Gehrmann, et~al.]{chowdhery2022palm}
Aakanksha Chowdhery, Sharan Narang, Jacob Devlin, Maarten Bosma, Gaurav Mishra,
  Adam Roberts, Paul Barham, Hyung~Won Chung, Charles Sutton, Sebastian
  Gehrmann, et~al.
\newblock Palm: Scaling language modeling with pathways.
\newblock \emph{arXiv preprint arXiv:2204.02311}, 2022.

\bibitem[Chung et~al.(2022)Chung, Hou, Longpre, Zoph, Tay, Fedus, Li, Wang,
  Dehghani, Brahma, et~al.]{chung2022scaling}
Hyung~Won Chung, Le~Hou, Shayne Longpre, Barret Zoph, Yi~Tay, William Fedus,
  Eric Li, Xuezhi Wang, Mostafa Dehghani, Siddhartha Brahma, et~al.
\newblock Scaling instruction-finetuned language models.
\newblock \emph{arXiv preprint arXiv:2210.11416}, 2022.

\bibitem[Dai et~al.(2023)Dai, Li, Li, Tiong, Zhao, Wang, Li, Fung, and
  Hoi]{instructblip}
Wenliang Dai, Junnan Li, Dongxu Li, Anthony Meng~Huat Tiong, Junqi Zhao,
  Weisheng Wang, Boyang Li, Pascale Fung, and Steven Hoi.
\newblock Instructblip: Towards general-purpose vision-language models with
  instruction tuning, 2023.

\bibitem[Gao et~al.(2023)Gao, Han, Zhang, Lin, Geng, Zhou, Zhang, Lu, He, Yue,
  et~al.]{gao2023llama}
Peng Gao, Jiaming Han, Renrui Zhang, Ziyi Lin, Shijie Geng, Aojun Zhou, Wei
  Zhang, Pan Lu, Conghui He, Xiangyu Yue, et~al.
\newblock Llama-adapter v2: Parameter-efficient visual instruction model.
\newblock \emph{arXiv preprint arXiv:2304.15010}, 2023.

\bibitem[Houlsby et~al.(2019)Houlsby, Giurgiu, Jastrzebski, Morrone,
  De~Laroussilhe, Gesmundo, Attariyan, and Gelly]{houlsby}
Neil Houlsby, Andrei Giurgiu, Stanislaw Jastrzebski, Bruna Morrone, Quentin
  De~Laroussilhe, Andrea Gesmundo, Mona Attariyan, and Sylvain Gelly.
\newblock Parameter-efficient transfer learning for nlp.
\newblock In \emph{International Conference on Machine Learning}, pp.\
  2790--2799. PMLR, 2019.

\bibitem[Hu et~al.(2021)Hu, Shen, Wallis, Allen-Zhu, Li, Wang, Wang, and
  Chen]{hu2021lora}
Edward~J Hu, Yelong Shen, Phillip Wallis, Zeyuan Allen-Zhu, Yuanzhi Li, Shean
  Wang, Lu~Wang, and Weizhu Chen.
\newblock Lora: Low-rank adaptation of large language models.
\newblock \emph{arXiv preprint arXiv:2106.09685}, 2021.

\bibitem[Lester et~al.(2021{\natexlab{a}})Lester, Al-Rfou, and
  Constant]{lester2021power}
Brian Lester, Rami Al-Rfou, and Noah Constant.
\newblock The power of scale for parameter-efficient prompt tuning.
\newblock \emph{arXiv preprint arXiv:2104.08691}, 2021{\natexlab{a}}.

\bibitem[Lester et~al.(2021{\natexlab{b}})Lester, Al{-}Rfou, and
  Constant]{prompt_tuning}
Brian Lester, Rami Al{-}Rfou, and Noah Constant.
\newblock The power of scale for parameter-efficient prompt tuning.
\newblock \emph{CoRR}, abs/2104.08691, 2021{\natexlab{b}}.
\newblock URL \url{https://arxiv.org/abs/2104.08691}.

\bibitem[Li et~al.(2023)Li, Li, Savarese, and Hoi]{li2023blip}
Junnan Li, Dongxu Li, Silvio Savarese, and Steven Hoi.
\newblock Blip-2: Bootstrapping language-image pre-training with frozen image
  encoders and large language models.
\newblock \emph{arXiv preprint arXiv:2301.12597}, 2023.

\bibitem[Li \& Liang(2021)Li and Liang]{prefix}
Xiang~Lisa Li and Percy Liang.
\newblock Prefix-tuning: Optimizing continuous prompts for generation.
\newblock \emph{CoRR}, abs/2101.00190, 2021.
\newblock URL \url{https://arxiv.org/abs/2101.00190}.

\bibitem[Lin et~al.(2014)Lin, Maire, Belongie, Hays, Perona, Ramanan,
  Doll{\'a}r, and Zitnick]{lin2014microsoft}
Tsung-Yi Lin, Michael Maire, Serge Belongie, James Hays, Pietro Perona, Deva
  Ramanan, Piotr Doll{\'a}r, and C~Lawrence Zitnick.
\newblock Microsoft coco: Common objects in context.
\newblock In \emph{Computer Vision--ECCV 2014: 13th European Conference,
  Zurich, Switzerland, September 6-12, 2014, Proceedings, Part V 13}, pp.\
  740--755. Springer, 2014.

\bibitem[Liu et~al.(2023)Liu, Li, Wu, and Lee]{liu2023visual}
Haotian Liu, Chunyuan Li, Qingyang Wu, and Yong~Jae Lee.
\newblock Visual instruction tuning.
\newblock \emph{arXiv preprint arXiv:2304.08485}, 2023.

\bibitem[Liu et~al.(2021)Liu, Ji, Fu, Tam, Du, Yang, and Tang]{liu2021p}
Xiao Liu, Kaixuan Ji, Yicheng Fu, Weng~Lam Tam, Zhengxiao Du, Zhilin Yang, and
  Jie Tang.
\newblock P-tuning v2: Prompt tuning can be comparable to fine-tuning
  universally across scales and tasks.
\newblock \emph{arXiv preprint arXiv:2110.07602}, 2021.

\bibitem[Longpre et~al.(2023)Longpre, Hou, Vu, Webson, Chung, Tay, Zhou, Le,
  Zoph, Wei, et~al.]{longpre2023flan}
Shayne Longpre, Le~Hou, Tu~Vu, Albert Webson, Hyung~Won Chung, Yi~Tay, Denny
  Zhou, Quoc~V Le, Barret Zoph, Jason Wei, et~al.
\newblock The flan collection: Designing data and methods for effective
  instruction tuning.
\newblock \emph{arXiv preprint arXiv:2301.13688}, 2023.

\bibitem[Mao et~al.(2021)Mao, Mathias, Hou, Almahairi, Ma, Han, Yih, and
  Khabsa]{mao2021unipelt}
Yuning Mao, Lambert Mathias, Rui Hou, Amjad Almahairi, Hao Ma, Jiawei Han,
  Wen-tau Yih, and Madian Khabsa.
\newblock Unipelt: A unified framework for parameter-efficient language model
  tuning.
\newblock \emph{arXiv preprint arXiv:2110.07577}, 2021.

\bibitem[Marino et~al.(2019)Marino, Rastegari, Farhadi, and
  Mottaghi]{marino2019ok}
Kenneth Marino, Mohammad Rastegari, Ali Farhadi, and Roozbeh Mottaghi.
\newblock Ok-vqa: A visual question answering benchmark requiring external
  knowledge.
\newblock In \emph{Proceedings of the IEEE/cvf conference on computer vision
  and pattern recognition}, pp.\  3195--3204, 2019.

\bibitem[Mathew et~al.(2021)Mathew, Karatzas, and Jawahar]{mathew2021docvqa}
Minesh Mathew, Dimosthenis Karatzas, and CV~Jawahar.
\newblock Docvqa: A dataset for vqa on document images.
\newblock In \emph{Proceedings of the IEEE/CVF winter conference on
  applications of computer vision}, pp.\  2200--2209, 2021.

\bibitem[Mishra et~al.(2022)Mishra, Khashabi, Baral, and
  Hajishirzi]{mishra2022cross}
Swaroop Mishra, Daniel Khashabi, Chitta Baral, and Hannaneh Hajishirzi.
\newblock Cross-task generalization via natural language crowdsourcing
  instructions.
\newblock In \emph{Proceedings of the 60th Annual Meeting of the Association
  for Computational Linguistics (Volume 1: Long Papers)}, pp.\  3470--3487,
  2022.

\bibitem[OpenAI(2023)]{OpenAI2023GPT4TR}
OpenAI.
\newblock Gpt-4 technical report.
\newblock \emph{ArXiv}, abs/2303.08774, 2023.

\bibitem[Ouyang et~al.(2022)Ouyang, Wu, Jiang, Almeida, Wainwright, Mishkin,
  Zhang, Agarwal, Slama, Ray, et~al.]{ouyang2022training}
Long Ouyang, Jeffrey Wu, Xu~Jiang, Diogo Almeida, Carroll Wainwright, Pamela
  Mishkin, Chong Zhang, Sandhini Agarwal, Katarina Slama, Alex Ray, et~al.
\newblock Training language models to follow instructions with human feedback.
\newblock \emph{Advances in Neural Information Processing Systems},
  35:\penalty0 27730--27744, 2022.

\bibitem[Pfeiffer et~al.(2020)Pfeiffer, R\"uckl\'{e}, Poth, Kamath, Vuli\'{c},
  Ruder, Cho, and Gurevych]{pfeiffer2020AdapterHub}
Jonas Pfeiffer, Andreas R\"uckl\'{e}, Clifton Poth, Aishwarya Kamath, Ivan
  Vuli\'{c}, Sebastian Ruder, Kyunghyun Cho, and Iryna Gurevych.
\newblock Adapterhub: A framework for adapting transformers.
\newblock In \emph{Proceedings of the 2020 Conference on Empirical Methods in
  Natural Language Processing (EMNLP 2020): Systems Demonstrations}, pp.\
  46--54, Online, 2020. Association for Computational Linguistics.
\newblock URL \url{https://www.aclweb.org/anthology/2020.emnlp-demos.7}.

\bibitem[Pfeiffer et~al.(2021)Pfeiffer, Kamath, R{\"u}ckl{\'e}, Cho, and
  Gurevych]{pfeiffer2021adapter}
Jonas Pfeiffer, Aishwarya Kamath, Andreas R{\"u}ckl{\'e}, Kyunghyun Cho, and
  Iryna Gurevych.
\newblock Adapterfusion: Non-destructive task composition for transfer
  learning.
\newblock In \emph{Proceedings of the 16th Conference of the European Chapter
  of the Association for Computational Linguistics: Main Volume}, pp.\
  487--503, 2021.

\bibitem[Raffel et~al.(2020)Raffel, Shazeer, Roberts, Lee, Narang, Matena,
  Zhou, Li, and Liu]{T5}
Colin Raffel, Noam Shazeer, Adam Roberts, Katherine Lee, Sharan Narang, Michael
  Matena, Yanqi Zhou, Wei Li, and Peter~J Liu.
\newblock Exploring the limits of transfer learning with a unified text-to-text
  transformer.
\newblock \emph{Journal of Machine Learning Research}, 21\penalty0
  (140):\penalty0 1--67, 2020.

\bibitem[Razdaibiedina et~al.(2023)Razdaibiedina, Mao, Hou, Khabsa, Lewis, Ba,
  and Almahairi]{razdaibiedina2023residual}
Anastasia Razdaibiedina, Yuning Mao, Rui Hou, Madian Khabsa, Mike Lewis, Jimmy
  Ba, and Amjad Almahairi.
\newblock Residual prompt tuning: Improving prompt tuning with residual
  reparameterization.
\newblock \emph{arXiv preprint arXiv:2305.03937}, 2023.

\bibitem[Sanh et~al.(2021)Sanh, Webson, Raffel, Bach, Sutawika, Alyafeai,
  Chaffin, Stiegler, Scao, Raja, et~al.]{sanh2021multitask}
Victor Sanh, Albert Webson, Colin Raffel, Stephen~H Bach, Lintang Sutawika,
  Zaid Alyafeai, Antoine Chaffin, Arnaud Stiegler, Teven~Le Scao, Arun Raja,
  et~al.
\newblock Multitask prompted training enables zero-shot task generalization.
\newblock \emph{arXiv preprint arXiv:2110.08207}, 2021.

\bibitem[Shazeer \& Stern(2018)Shazeer and Stern]{shazeer2018adafactor}
Noam Shazeer and Mitchell Stern.
\newblock Adafactor: Adaptive learning rates with sublinear memory cost.
\newblock In \emph{International Conference on Machine Learning}, pp.\
  4596--4604. PMLR, 2018.

\bibitem[Sidorov et~al.(2020)Sidorov, Hu, Rohrbach, and
  Singh]{sidorov2020textcaps}
Oleksii Sidorov, Ronghang Hu, Marcus Rohrbach, and Amanpreet Singh.
\newblock Textcaps: a dataset for image captioning with reading comprehension.
\newblock In \emph{Computer Vision--ECCV 2020: 16th European Conference,
  Glasgow, UK, August 23--28, 2020, Proceedings, Part II 16}, pp.\  742--758.
  Springer, 2020.

\bibitem[Singh et~al.(2019)Singh, Natarjan, Shah, Jiang, Chen, Parikh, and
  Rohrbach]{singh2019towards}
Amanpreet Singh, Vivek Natarjan, Meet Shah, Yu~Jiang, Xinlei Chen, Devi Parikh,
  and Marcus Rohrbach.
\newblock Towards vqa models that can read.
\newblock In \emph{Proceedings of the IEEE Conference on Computer Vision and
  Pattern Recognition}, pp.\  8317--8326, 2019.

\bibitem[Su et~al.(2022)Su, Wang, Qin, Chan, Lin, Wang, Wen, Liu, Li, Li,
  et~al.]{su2022transferability}
Yusheng Su, Xiaozhi Wang, Yujia Qin, Chi-Min Chan, Yankai Lin, Huadong Wang,
  Kaiyue Wen, Zhiyuan Liu, Peng Li, Juanzi Li, et~al.
\newblock On transferability of prompt tuning for natural language processing.
\newblock In \emph{Proceedings of the 2022 Conference of the North American
  Chapter of the Association for Computational Linguistics: Human Language
  Technologies}, pp.\  3949--3969, 2022.

\bibitem[Touvron et~al.(2023)Touvron, Martin, Stone, Albert, Almahairi, Babaei,
  Bashlykov, Batra, Bhargava, Bhosale, et~al.]{llama2}
Hugo Touvron, Louis Martin, Kevin Stone, Peter Albert, Amjad Almahairi, Yasmine
  Babaei, Nikolay Bashlykov, Soumya Batra, Prajjwal Bhargava, Shruti Bhosale,
  et~al.
\newblock Llama 2: Open foundation and fine-tuned chat models.
\newblock \emph{arXiv preprint arXiv:2307.09288}, 2023.

\bibitem[Vaswani et~al.(2017)Vaswani, Shazeer, Parmar, Uszkoreit, Jones, Gomez,
  Kaiser, and Polosukhin]{NIPS2017_3f5ee243}
Ashish Vaswani, Noam Shazeer, Niki Parmar, Jakob Uszkoreit, Llion Jones,
  Aidan~N Gomez, \L~ukasz Kaiser, and Illia Polosukhin.
\newblock Attention is all you need.
\newblock In I.~Guyon, U.~Von Luxburg, S.~Bengio, H.~Wallach, R.~Fergus,
  S.~Vishwanathan, and R.~Garnett (eds.), \emph{Advances in Neural Information
  Processing Systems}, volume~30. Curran Associates, Inc., 2017.
\newblock URL
  \url{https://proceedings.neurips.cc/paper_files/paper/2017/file/3f5ee243547dee91fbd053c1c4a845aa-Paper.pdf}.

\bibitem[Vedantam et~al.(2015)Vedantam, Lawrence~Zitnick, and
  Parikh]{vedantam2015cider}
Ramakrishna Vedantam, C~Lawrence~Zitnick, and Devi Parikh.
\newblock Cider: Consensus-based image description evaluation.
\newblock In \emph{Proceedings of the IEEE conference on computer vision and
  pattern recognition}, pp.\  4566--4575, 2015.

\bibitem[Wang et~al.(2019)Wang, Singh, Michael, Hill, Levy, and
  Bowman]{wang2019glue}
Alex Wang, Amanpreet Singh, Julian Michael, Felix Hill, Omer Levy, and Samuel~R
  Bowman.
\newblock {GLUE}: A multi-task benchmark and analysis platform for natural
  language understanding.
\newblock 2019.

\bibitem[Wang et~al.(2022{\natexlab{a}})Wang, Bao, Dong, Bjorck, Peng, Liu,
  Aggarwal, Mohammed, Singhal, Som, et~al.]{wang2022image}
Wenhui Wang, Hangbo Bao, Li~Dong, Johan Bjorck, Zhiliang Peng, Qiang Liu, Kriti
  Aggarwal, Owais~Khan Mohammed, Saksham Singhal, Subhojit Som, et~al.
\newblock Image as a foreign language: Beit pretraining for all vision and
  vision-language tasks.
\newblock \emph{arXiv preprint arXiv:2208.10442}, 2022{\natexlab{a}}.

\bibitem[Wang et~al.(2022{\natexlab{b}})Wang, Kordi, Mishra, Liu, Smith,
  Khashabi, and Hajishirzi]{wang2022self}
Yizhong Wang, Yeganeh Kordi, Swaroop Mishra, Alisa Liu, Noah~A Smith, Daniel
  Khashabi, and Hannaneh Hajishirzi.
\newblock Self-instruct: Aligning language model with self generated
  instructions.
\newblock \emph{arXiv preprint arXiv:2212.10560}, 2022{\natexlab{b}}.

\bibitem[Wang et~al.(2022{\natexlab{c}})Wang, Panda, Karlinsky, Feris, Sun, and
  Kim]{wang2022multitask}
Zhen Wang, Rameswar Panda, Leonid Karlinsky, Rogerio Feris, Huan Sun, and Yoon
  Kim.
\newblock Multitask prompt tuning enables parameter-efficient transfer
  learning.
\newblock In \emph{The Eleventh International Conference on Learning
  Representations}, 2022{\natexlab{c}}.

\bibitem[Wei et~al.(2021)Wei, Bosma, Zhao, Guu, Yu, Lester, Du, Dai, and
  Le]{wei2021finetuned}
Jason Wei, Maarten Bosma, Vincent~Y Zhao, Kelvin Guu, Adams~Wei Yu, Brian
  Lester, Nan Du, Andrew~M Dai, and Quoc~V Le.
\newblock Finetuned language models are zero-shot learners.
\newblock \emph{arXiv preprint arXiv:2109.01652}, 2021.

\bibitem[Yang et~al.(2022)Yang, Lin, Yang, Wang, Zhou, and
  Yang]{yang2022prompt}
Hao Yang, Junyang Lin, An~Yang, Peng Wang, Chang Zhou, and Hongxia Yang.
\newblock Prompt tuning for generative multimodal pretrained models.
\newblock \emph{arXiv preprint arXiv:2208.02532}, 2022.

\bibitem[Zhang et~al.(2023)Zhang, Han, Zhou, Hu, Yan, Lu, Li, Gao, and
  Qiao]{zhang2023llama}
Renrui Zhang, Jiaming Han, Aojun Zhou, Xiangfei Hu, Shilin Yan, Pan Lu,
  Hongsheng Li, Peng Gao, and Yu~Qiao.
\newblock Llama-adapter: Efficient fine-tuning of language models with
  zero-init attention.
\newblock \emph{arXiv preprint arXiv:2303.16199}, 2023.

\bibitem[Zhong et~al.(2022)Zhong, Ding, Liu, Du, and Tao]{zhong2022panda}
Qihuang Zhong, Liang Ding, Juhua Liu, Bo~Du, and Dacheng Tao.
\newblock Panda: Prompt transfer meets knowledge distillation for efficient
  model adaptation.
\newblock \emph{arXiv preprint arXiv:2208.10160}, 2022.

\end{thebibliography}
\bibliographystyle{iclr2024_conference}

% \appendix
% \section{Appendix}

\end{document}